\documentclass[10pt,twocolumn,letterpaper]{article}

\usepackage{cvpr}
\usepackage{times}
\usepackage{epsfig}
\usepackage{graphicx}
\usepackage{amsmath}
\usepackage{amssymb}
\usepackage{multirow}
\usepackage{enumitem}
\usepackage{algorithm}
\usepackage{algpseudocode}% http://ctan.org/pkg/algorithmicx
\usepackage{subfigure}
\usepackage{booktabs}
\usepackage[dvipsnames]{xcolor}
\usepackage{stfloats}
\newfloat{algorithm}{t}{lop}
\usepackage{color, colortbl}
\definecolor{Gray}{gray}{0.95}
\newcommand{\beginsupplement}{%
        \setcounter{table}{0}
        \renewcommand{\thetable}{A\arabic{table}}%
        \setcounter{figure}{0}
        \renewcommand{\thefigure}{A\arabic{figure}}%
        \setcounter{section}{0}
        \renewcommand{\thesection}{\Alph{section}} 
     }

\usepackage[pagebackref=true,breaklinks=true,letterpaper=true,colorlinks,bookmarks=false]{hyperref}

\cvprfinalcopy % *** Uncomment this line for the final submission

\def\cvprPaperID{6287} % *** Enter the CVPR Paper ID here

\newcommand{\fullname}{\textbf{F}rontier \textbf{A}ware \textbf{S}earch with back\textbf{T}racking}
\newcommand{\decoder}{\textsc{Fast Navigator}}
\newcommand{\short}{\textsc{Fast}}

\newcommand{\nop}[1]{}

\definecolor{pink}{cmyk}{0, 0.7808, 0.4429, 0.1412}

\ifcvprfinal\fi
\begin{document}

%%%%%%%%% TITLE
\title{Tactical Rewind: Self-Correction via Backtracking\\ in Vision-and-Language Navigation}

\author{Liyiming Ke$^1$\thanks{Work done partially as an intern at MSR}\hspace{2em} Xiujun Li\textsuperscript{1,2}\hspace{2em} Yonatan Bisk\textsuperscript{1}\hspace{2em} Ari Holtzman\textsuperscript{$1$}\hspace{2em} Zhe Gan\textsuperscript{2}\\
Jingjing Liu\textsuperscript{2}\hspace{2em} Jianfeng Gao\textsuperscript{2}\hspace{2em} Yejin Choi\textsuperscript{1,3}\hspace{2em} Siddhartha Srinivasa\textsuperscript{1}\\
%\textsuperscript{$\star$}The Ohio State University\quad\quad 
\textsuperscript{1}Paul G. Allen School of Computer Science \& Engineering, University of Washington\\
\textsuperscript{2}Microsoft Research AI\quad\quad\quad\quad \textsuperscript{3}Allen Institute for Artificial Intelligence\\
{\tt\small \{kayke, xiujun, ybisk, ahai, yejin, siddh\}@cs.washington.edu}\\
{\tt\small \{xiul, zhgan, jingjl, jfgao\}@microsoft.com}
}

% Recommended Authors: Liyiming Ke, Xiujun Li, Yonatan Bisk, Ari Holtzman, Zhe Gan, Jingjing Liu, Jianfeng Gao, Yejin Choi and Siddhartha Srinivasa

%\and

\maketitle
\thispagestyle{empty}

%%%%%%%%% ABSTRACT
\begin{abstract}
We present the Frontier Aware Search with backTracking (\short{}) Navigator, a general framework for action decoding, that achieves state-of-the-art results on the Room-to-Room (R2R) Vision-and-Language navigation challenge of Anderson et.\ al.\ (2018). Given a natural language instruction and photo-realistic image views of a previously unseen environment, the agent was tasked with navigating from source to target location as quickly as possible. While all current approaches make local action decisions or score entire trajectories using beam search, ours balances local and global signals when exploring an unobserved environment. Importantly, this lets us act greedily but use global signals to backtrack when necessary. Applying \short{} framework to existing state-of-the-art models achieved a 17\% relative gain, an absolute 6\% gain on Success rate weighted by Path Length (SPL).\footnote{The code is available at \url{https://github.com/Kelym/FAST}.}

%We present \decoder{}, a general framework for action decoding, which yields new state-of-the-art results on the recently released Room-to-Room (R2R) Vision-and-Language navigation challenge of Anderson et.\ al.\ (2018). R2R requires an agent to follow a novel natural language instruction in a previously unseen environment. In R2R the agent attempts to navigate to the stated target location as quickly and accurately as possible. While other approaches make local action decisions or score entire trajectories with beam search, our framework seamlessly balances local and global signals when exploring the environment. Importantly, this allows us to act greedily, but use global signals to backtrack when necessary. Our \short{} framework can be applied to improve all existing models, and as such we see a 17\% relative gain over the previous state-of-the-art on unseen environments (an absolute gain of 6 SPL).
\end{abstract}

\section{Introduction}
\label{sec:intro}

When reading an instruction (e.g. ``\textit{Exit the bathroom, take the second door on your right, pass the sofa and stop at the top of the stairs .}''), a person builds a mental map of how to arrive at a specific location. This map can include landmarks, such as the second door, and markers  such as reaching the top of the stairs. Training an embodied agent to accomplish such a task with access to only ego-centric vision and individually supervised actions requires building rich multi-modal representations from limited data~\cite{anderson2018vision}.

Most current approaches to Vision-and-Language Navigation (VLN) formulate the task to use the seq2seq (or encoder-decoder) framework~\cite{sutskever2014sequence}, where language and vision are encoded as input and an optimal action sequence is decoded as output. Several subsequent architectures also use this framing; however, they augment it with important advances in attention mechanisms, global scoring, and beam search~\cite{anderson2018vision,ma2019self,fried2018speaker}. %All approaches to VLN correctly treat the task as a search problem, but vacillate between either decoding greedily with local information or globally with beam search.

%\begin{figure}[t]
%\centering
%\includegraphics[width=\linewidth]{imgs/teaser.png}
%\caption{In Vision-and-Language Navigation, agent can get lost once it deviates from the instructed path. \decoder{} remedies this by allowing ``tactical rewinds" when combined local and global information suggests that the current node is a dead-end and moving the agent to a better position. }
%\label{fig:method}
%\end{figure}

%\K{the paragraph is in revision. See the new paragraph after it.}
%With this formulation, comes the problem of \textit{exposure bias}~\cite{ranzato2015sequence}, wherein a model that has only been trained to predict one-step into the future based on ground-truth sequences is incapable of performing accurately using its own self-generated sequences. In particular, exposure bias often emerges when locally-optimal decisions lead to strange partial sequences and in turn extremely sub-optimal complete sequence (Figure~\ref{fig:method}). Beam search attempts to remedy this by first generating $B$ candidate action sequences using local information, then re-scoring these $B$ complete candidate action sequences using global information. \YB{shorten paragraph, add discussion about student sampling, and how this is a technique people use to overcome exposure in s2s}
%Sequence models in general are known to suffer from the \textit{exposure bias} problem

\begin{figure}[t!]
\centering 
\subfigure[SoTA Beam Search]{
\label{fig:sota_graph} 
\includegraphics[width=0.42\columnwidth]{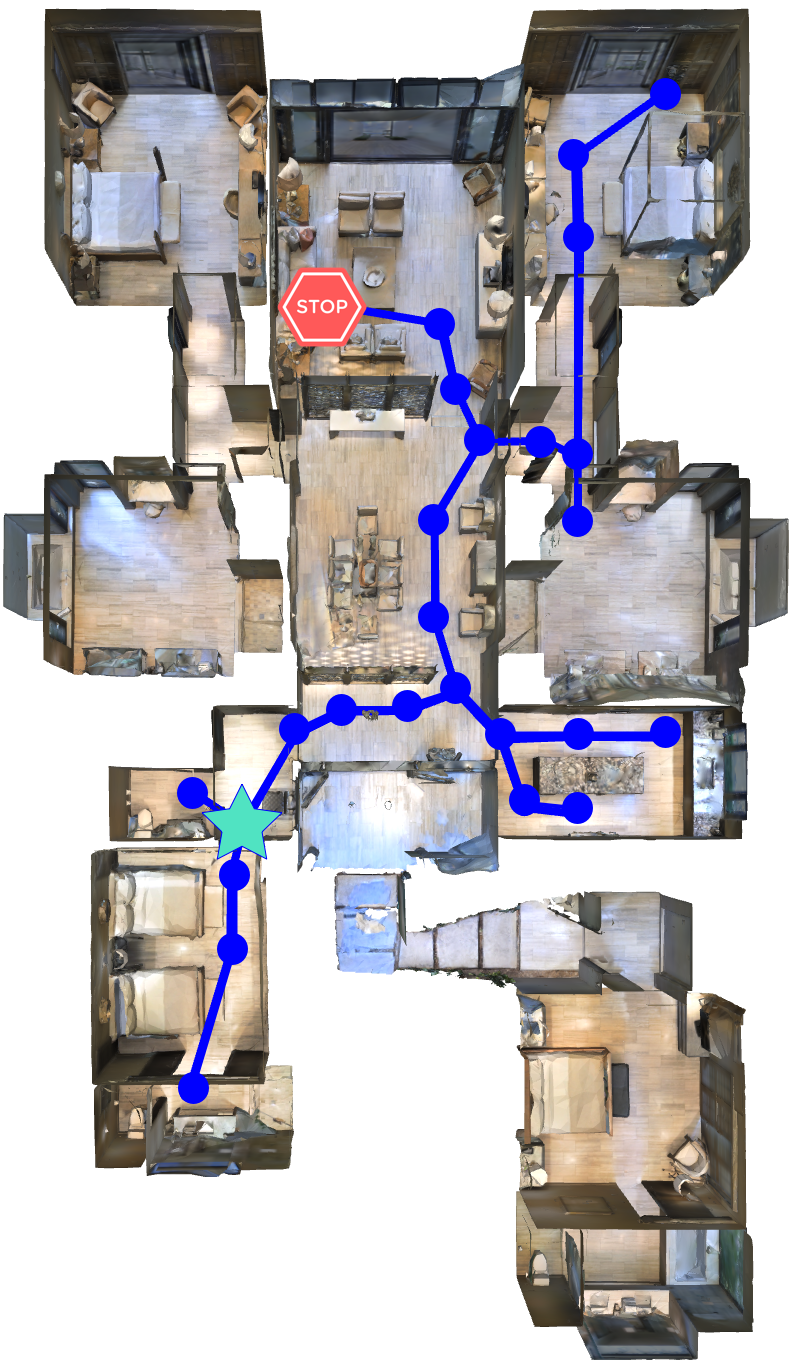}
}
\hspace{20pt}
\subfigure[\decoder{}]{
\label{fig:our_agent_graph} 
\includegraphics[width=0.42\columnwidth]{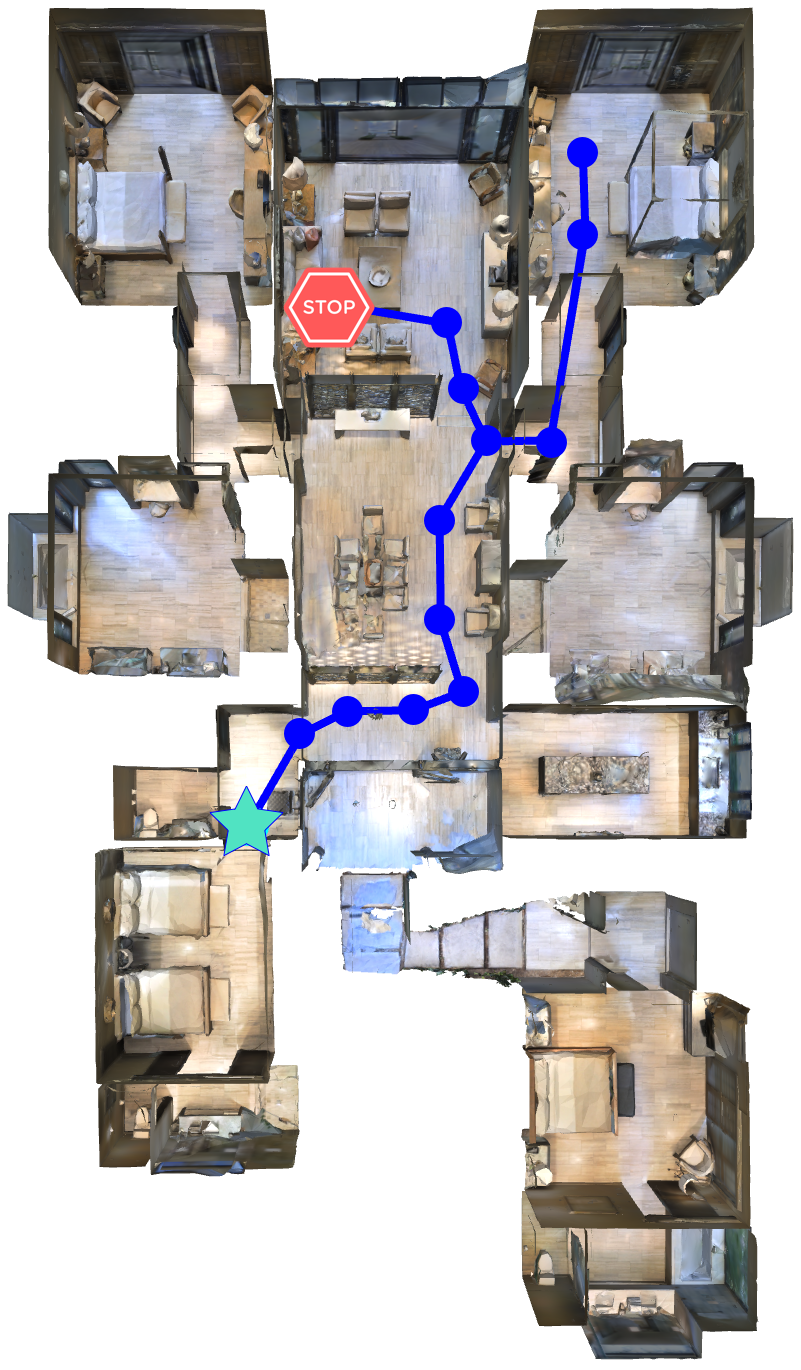}
}
\caption{
Top-down view of the trajectory graphs for beam search and \short{}. \textcolor{Turquoise}{Blue Star} is the start and \textcolor{red}{Red Stop} is the target.} 
\label{fig:trajectory_graphs} 
\end{figure}

Inherent to the seq2seq formulation is the problem of \textit{exposure bias} \cite{ranzato2015sequence}: a model that has been trained to predict one-step into the future given the ground-truth sequence cannot perform accurately given its self-generated sequence. Previous work with seq2seq models attempted to address this using student forcing and beam search. 

\textit{Student forcing} exposes a model to its own generated sequence during training, teaching the agent how to recover. However, once the agent has deviated from the correct path, the original instruction no longer applies. The Supplementary Materials (\S\ref{sec:app_examples}) show that student forcing cannot solve the exposure bias problem, causing the confused agent to fall into loops. 

%Though empirically alleviate the exposure bias, we show in result section that this cannot solve the exposure problem completely and often traverse back-and-force between pair of nodes when it get confused. 

\textit{Beam search}, at the other extreme, collects multiple global trajectories to score and incurs a cost proportional to the number of trajectories, which can be prohibitively high. This approach runs counter to the goal of building an agent that can efficiently navigate an environment: No one would likely deploy a household robot that re-navigates an entire house 100 times\footnote{This is calculated based on the length of \textsc{Speaker-Follower} agent paths and human paths on the R2R dataset.} before executing each command, even if it ultimately arrives at the correct location. The top performing systems on the VLN leaderboard\footnote{https://evalai.cloudcv.org/web/challenges/challenge-page/97/leaderboard/270} all require broad exploration that yields long trajectories, causing poor SPL performance (\textbf{S}uccess weighted by \textbf{P}ath \textbf{L}ength~\cite{anderson2018evaluation}). 

To alleviate the issues of exposure bias and expensive, inefficient beam-search decoding, we propose the \fullname (\decoder{}). This framework lets agents \textit{compare partial paths of different lengths} based on local and global information and then backtrack if it discerns a mistake. Figure~\ref{fig:trajectory_graphs} shows trajectory graphs created by the current published state-of-the-art (SoTA) agent using beam search versus our own. 
%In contrast with prior work, our method does not need to expand every candidate path before we can stop the algorithm and propose. 

Our method is a form of asynchronous search, which combines global and local knowledge to score and compare partial trajectories of different lengths. We evaluate our progress to the goal by modeling how closely our previous actions align with the given text instructions. To achieve this, we use a \textit{fusion} function, which converts local action knowledge and history into an estimated score of progress. This score determines which local action to take and whether the agent should backtrack. 
%The primary contribution of our work is to marry neural decoding with traditional approaches to search. 
This insight yields significant gains on evaluation metrics relative to existing models. The primary contributions of our work are:
\begin{itemize}[noitemsep,topsep=1pt,leftmargin=*]
    \item A method to alleviate the exposure bias of action decoding and expensiveness of beam search. 
    \item An algorithm that makes use of asynchronous search with neural decoding.
    \item An extensible framework that can be applied to existing models to achieve significant gains on SPL.
\end{itemize}

\section{Method}
\label{sec:method}

The VLN challenge requires an agent to carry out a natural language instruction in photo-realistic environments. The agent takes an input instruction $\mathcal{X}$, which contains several sentences describing a desired trajectory. At each step $t$, the agent observes its surroundings $\mathcal{V}_t$. Because the agent can look around for 360 degrees, $\mathcal{V}_t$ is in fact a set of $K=36$ different views. We denote each view as $\mathcal{V}^k_t$. Using this multimodal input, the agent is trained to execute a sequence of actions ${a_1, a_2, ...., a_T} \in \mathcal{A}$ to reach a desired location. Consistent with recent work~\cite{ma2019self,fried2018speaker}, we use a panoramic action space, where each action corresponds to moving towards one of the $K$ views, instead of R2R's original primitive action space (i.e, \textit{left}, \textit{right}, etc.) \cite{anderson2018vision,wang2018look}. In addition, this formulation includes a $stop$ action to indicate that the agent has reached its goal.

%VLN asks an agent to carry out an instruction in a photo-realistic environment. In each episode, the agent is given a natural language instruction $X = {x_1, x_2, ..., x_L}$ with $L$ words. At each step, the agent observes a set of images of its surroundings $v_t = {v_{t,1}, v_{t,2}, ..., v_{t,K}}$, where $v_{t,i}$ is the image feature of direction $i$, $K$ is the number of navigable directions. The agent is trained to execute a sequence of actions ${a_1, a_2, ...., a_T} \in \mathcal{A}$ to reach the target location. The literature takes two approaches to defining the action space $\mathcal{A}$. The original work~\cite{anderson2018vision,wang2018look} defines a minimal set of six unique actions $\{left, right, up, down, forward$ and $stop\}$, while newer approaches define a panoramic action space~\cite{selfaware2018,fried2018speaker}, which consists of 36 actions and one $stop$ action. In our work, we follow the example of top performers and utilize a panoramic action space.

\begin{figure}[ht!]
\centering
\hspace{0pt}Greedy \hspace{33pt} \short{} \hspace{32pt} Beam Search\hspace{20pt}\\
\includegraphics[width=0.9\columnwidth]{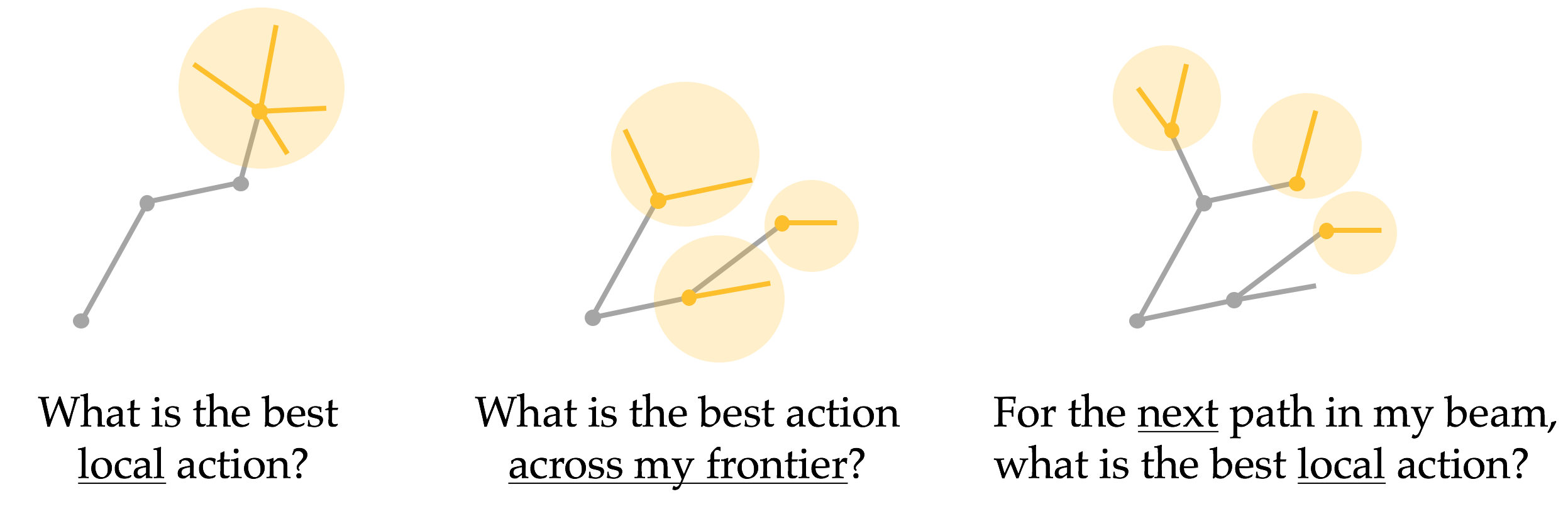}
\caption{All VLN agents are performing a search. The orange areas highlight the frontier for different navigation methods. 
}
\label{fig:frontier}
\end{figure}

\subsection{Learning Signals}
\label{sec:signals}
Key to progress in visual navigation is that \textit{all} VLN approaches performs a search (Figure \ref{fig:frontier}). Current work often goes toward two extremes: using only local information, e.g. greedy decoding, or fully sweeping multiple paths simultaneously, e.g. beam search. To build an agent that can navigate an environment successfully and efficiently, we leverage both \textbf{\textit{local}} and \textbf{\textit{global}} information, letting the agent make a local decision while remaining aware of its global progress and efficiently backtracking when the agent discerns a mistake. Inspired by previous work~\cite{fried2018speaker,ma2019self}, our work uses three learning signals:
%Our work integrates three signals from two previous papers: \textsc{Speaker-Follower}~\cite{fried2018speaker} and Self-Monitoring Navigation Agent (\textsc{SMNA})~\cite{selfaware2018}. 

\textbf{\textsc{Logit} $l_t$}: local distribution over action. The logit of the action chosen at time $t$ is denoted $l_t$. Specifically, the original language instruction is encoded via LSTM. Another LSTM acts as a decoder, using attention mechanism to generate logits over actions. At each time step $t$ of decoding, logits are calculated by taking the dot product of the decoder's hidden state and each candidate action $a_t^i$. 
%and uses attention mechanism to score each candidate action $a_t^i$ using a logit $l_t^i$.  %An logit for action $a_t$ is denoted as $l_t$. At each timestep there are candidates actions $a_t^i$, each assigned one logit $l_t^i$. Choosing the action $a_t^k$   Each of the $K$ views a \K{one logit is assigned to a pool of candidates  at each step $t$. The action taken $a_t$ is assigned logit $l_t$. we might want to clarify this?
%Specifically, an LSTM is used to encode the original language instruction into a sequence of hidden states, and the dot product of the hidden state at time $t$ with each action $a_t^i$ produces the logit. Denote the action chosen at time $t$ to have logit $l_t$.   

\textbf{\textsc{PM} $p_t^{pm}$}: global progress monitor. It tracks how much of an instruction has been completed ~\cite{ma2019self}. Formally, the model takes as input the (decoder) LSTM's current cell state, $c_t$, previous hidden state, $h_{t-1}$, visual inputs, $\mathcal{V}_t$, and attention over language embeddings, $\alpha_t$ to compute a score $p_t^{pm}$. The score ranges between [-1,1], indicating the agent's normalized progress. Training this indicator regularizes attention alignments, helping the model learn language-to-vision correspondences that it can use to compare multiple trajectories.

\textbf{\textsc{Speaker} $\mathcal{S}$}: global scoring. Given a sequence of visual observations and actions, we train a seq2seq captioning model as a ``speaker"~\cite{fried2018speaker} to produce a textual description. Doing so provides two benefits: (1) the new speaker can automatically annotate new trajectories in the environment with the synthetic instructions, and (2) the speaker can score the likelihood that a given trajectory will correspond to the original instruction.
%Important for our work, it introduces a new global scoring mechanism.
%\hfill \textbf{New signal}: $\mathcal{S}$

%This work introduces a progress monitor which tells the model how far along it is in completing the instruction. More formally, the model uses the LSTM's current content $c_t$, previous hidden state $h_{t-1}$, visual inputs $\mathcal{V}_t$, and attention over the language embeddings $\alpha_t$ to compute a score $p_t^{pm}$ that ranges [-1,1] and indicates the agent's normalized progress, the higher the better. Training this indicator appears to provide a regularizing effect on the attention alignments, helping the model learn language-to-vision correspondences. Secondarily, it can be used to compare multiple trajectories.\hfill \textbf{New signal}: $p_t^{pm}$

\begin{figure*}[t!]
\centering 
\subfigure[Instructions and visual observations are encoded as hidden vectors defining multiple paths through the world. These vectors can then be accumulated to score a sequence of actions.
%\XL{try to add Speaker S into the graph.}\K{@xj, I don't think we should add speaker to here or 4b? coz we are illustrating scoring for "partial trajectories" and speaker only applies to completed trajectories}
]
{
\label{fig:procedure} 
\includegraphics[width=0.465\textwidth]{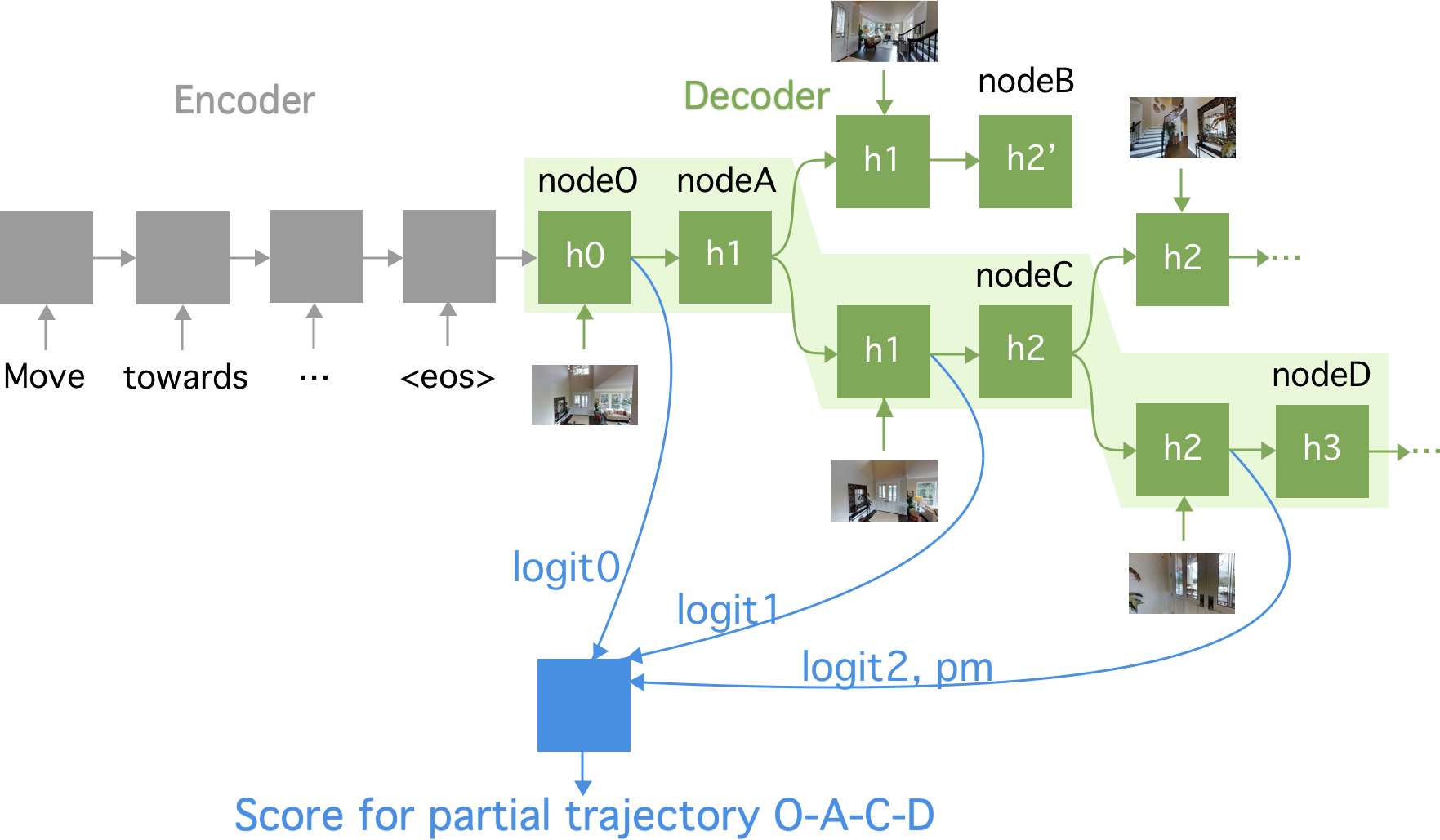}
}
\hspace{20pt}
\subfigure[At each time step, the predicted action sequence and visual observation are fed into an attention module with the encoded instruction, to produce both the logits for the next actions and a progress monitor score.]
{
\label{fig:smna_logits} 
\includegraphics[width=0.465\textwidth]{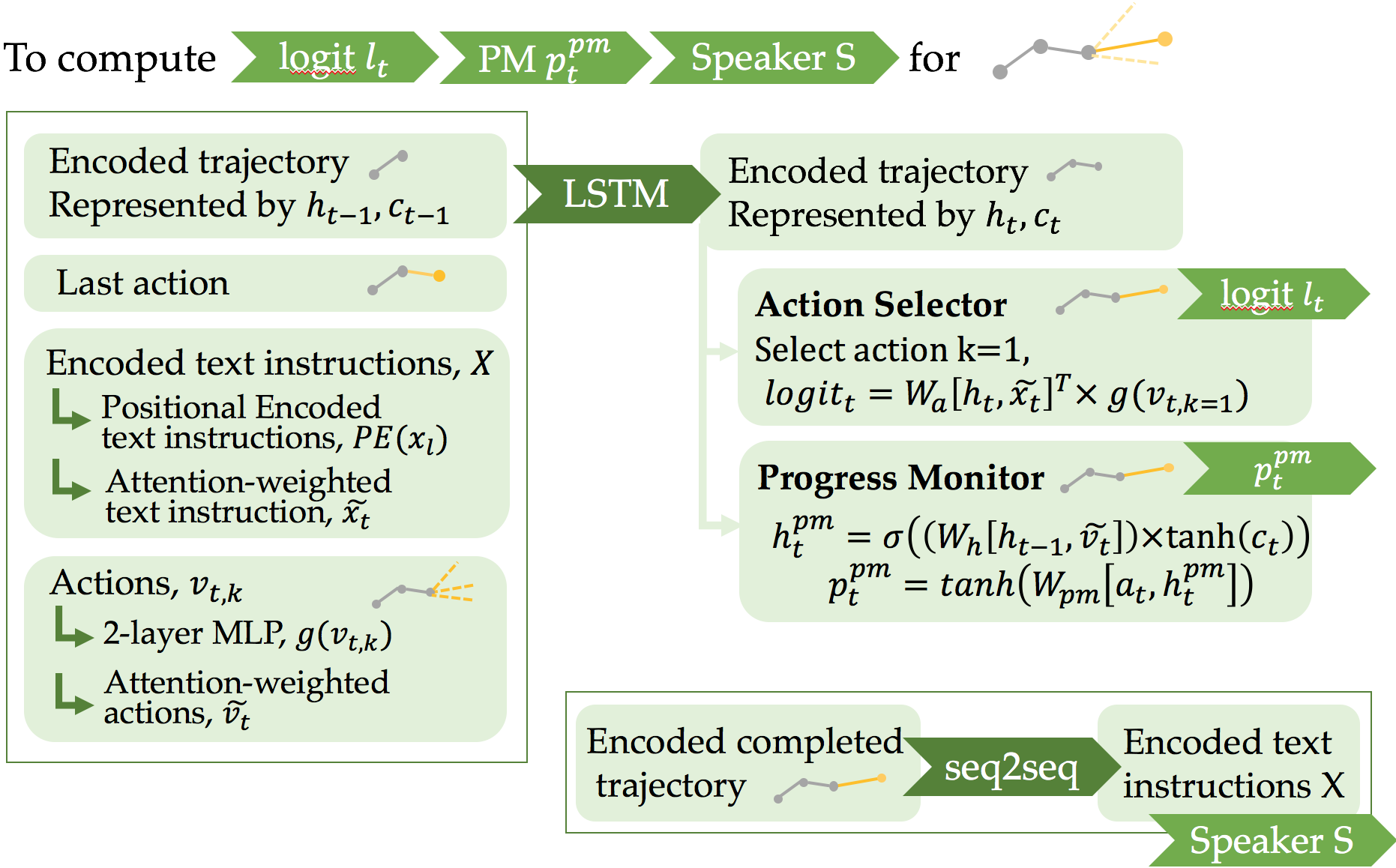}
}
\caption{(a). How the three signals are extracted from the partial trajectory in a seq2seq VLN framework; (b). How to compute the three signals.} 
\label{fig:signals} 
\end{figure*}

\subsection{Framework}
%\XL{re-structuring this section, to make it more clear.}
We now introduce an extendible framework\footnote{Figure~\ref{fig:procedure} shows an example of integrating the three signals in a seq2seq framework.} that integrates the preceding three signals ($l_t$, $p_t^{pm}$, $\mathcal{S}$)\footnote{Figure~\ref{fig:smna_logits} shows how to compute the three signals.} and to train new indicators, equipping an agent to answer:
%allow the agent at any point to answer:
\begin{enumerate}
    \itemsep-0.2em 
    \item Should we backtrack? 
    \item Where should we backtrack to?
    \item Which visited node is most likely to be the goal? 
    \item When does it terminate this search?
\end{enumerate}

%but previous approaches have not explicitly analyzed them. 
These questions pertain to all existing approaches in navigation task. In particular, greedy approaches never backtrack and do not compare partial trajectories. Global beam search techniques always backtrack but can waste efforts. By taking a more principled approach to modeling navigation as graph traversal, our framework permits nuanced and adaptive answers to each of these questions. 

%We formulate navigation as graph traversal. 
For navigation, the graph is defined by a series of locations in the environment, called nodes. For each task, the agent is placed at a starting node, and the agent's movement in the house creates a trajectory comprised of a sequence of $<$\textit{node} $u$, \textit{action} $a>$ pairs. We denote a \emph{partial} trajectory up to time $t$ as $\tau_t$, or the set of physical locations visited and the action taken at each point: 
\begin{equation}
\tau_t = \{(u_i, a_i)\}^t_{i=1}
\end{equation}

For any partial trajectory, the last action is \textit{proposed} and evaluated, but not \textit{executed}. Instead, the model chooses whether to expand a partial trajectory or execute a \textit{stop} action to complete the trajectory. Importantly, this means that every node the agent visited can serve as a possible final destination. The agent moves in the environment by choosing to extend a partial trajectory: it does this by moving to the last node of the partial trajectory and executing its last action to arrive at a new node. The agent then realizes the actions available at the new node and collects them to build a set of new partial trajectories.

%Note that every node that an agent has visited can be the final destination, as the last action of a completed trajectory is the \textit{stop} action. For any partial trajectory, the last action is \textit{proposed} but never \textit{executed}.
%\JG{We can point out explicitly that this is the difference between partial trajectories and completed trajectories. Note that every node that an agent has visited could be a final destination.}

At each time step, the agent must (1) access the set of partial trajectories it has not expanded, (2) access the completed trajectories that might constitute the candidate path, (3) calculate the accumulated cost of partial trajectories and the expected gain of its proposed action, and (4) compares all partial trajectories. 

To do so, we maintain two priority queues: a \textit{frontier queue}, $\mathcal{Q}_F$, for partial trajectories, and  a \textit{global candidate queue}, $\mathcal{Q}_C$, for completed trajectories. These queues are sorted by local $\mathcal{L}$ and global $\mathcal{G}$ scores, respectively. $\mathcal{L}$ scores the quality of all partial trajectories with their proposed actions and maintains their order in $\mathcal{Q}_F$; $\mathcal{G}$ scores the quality of completed trajectories and maintains the order in $\mathcal{Q}_C$. 

In \S\ref{sec:ablate_feats}, we explore alternative formulas for $\mathcal{L}$ and $\mathcal{G}$. For example, we define $\mathcal{L}$ and $\mathcal{G}$ using the signals described in \S\ref{sec:signals} and a function, $f$, that is implemented as a neural network.
\begin{eqnarray}
    \mathcal{L} && \leftarrow \Sigma_{0\rightarrow t}~l_i\\
    \mathcal{G} && \leftarrow f(\mathcal{S}, p_t^{pm}, \Sigma_{0\rightarrow t}~l_i, ... )
\end{eqnarray}

% where $f$ is parameterized by a neural network.  
% In practice, both $\mathcal{L}$ and $\mathcal{G}$ are trainable functions which can incorporate multiple signals. we define them in terms of the signals described in \S\ref{sec:signals}:

%\JG{We need to explain why PM and Spearker are not used in $\mathcal{L}$, and LOGIT not used in $\mathcal{G}$. Is this due to the performance or the difference between partial and completed trajectories (e.g., PM and Speaker are applicable to a trajectory only after its last action is executed)?}\XL{In \S\ref{sec:signals}, for each signal, I have one sentence to say it is local/global.}\K{S cannot serve as local information, because it can only evaluate completed trajectory; we experimented with using PM in local scorer in our experiment section. We didn't write it here.}

To allow the agent to efficiently navigate and follow the instruction, we use an approximation of the D* search. \short{} expands its optimal partial trajectory until it decides to backtrack (\textbf{Q1}). It decides on where to backtrack (\textbf{Q2}) by ranking all partial trajectories. To propose the final goal location (\textbf{Q3 \& Q4}), the agent ranks the completed global trajectories in candidate queue $\mathcal{Q}_C$. We explore these questions in more detail below. %The general architecture and feature computations are shown in Figure \ref{fig:signals}.

\paragraph{Q1: Should we backtrack?}
When an agent makes a mistake or gets lost, backtracking lets it move to a more promising partial trajectory; however, retracing steps increases the length of the final path. To determine when it is worth incurring this cost, we proposed two simple strategies: \textit{explore} and \textit{exploit}. 
\begin{enumerate}[noitemsep,topsep=2pt,leftmargin=*]
    \item \textit{Explore} always backtracks to the most promising partial trajectory. This approach resembles beam search, but, rather than simply moving to the next partial trajectory in the beam, the agent computes the most promising node to backtrack to (\textbf{Q2}).
    \item \textit{Exploit}, in contrast, commits to the current partial trajectory, always executing the best action available at the agent's current location. This approach resembles greedy decoding, except that the agent backtracks when it is confused (i.e, when the best local action causes the agent to revisit a node, creating a loop; see the SMNA examples in Supplementary Materials \S\ref{sec:supp_examples}). 
\end{enumerate}

\begin{figure*}[t!]
\centering 
\subfigure[Both local $\mathcal{L}$ and global $\mathcal{G}$ scores can be trained to condition on arbitrary information. Here, we show the fusion of historical logits and progress monitor information into a single score.]{
\label{fig:fast-fusion} 
\includegraphics[width=0.46\textwidth]{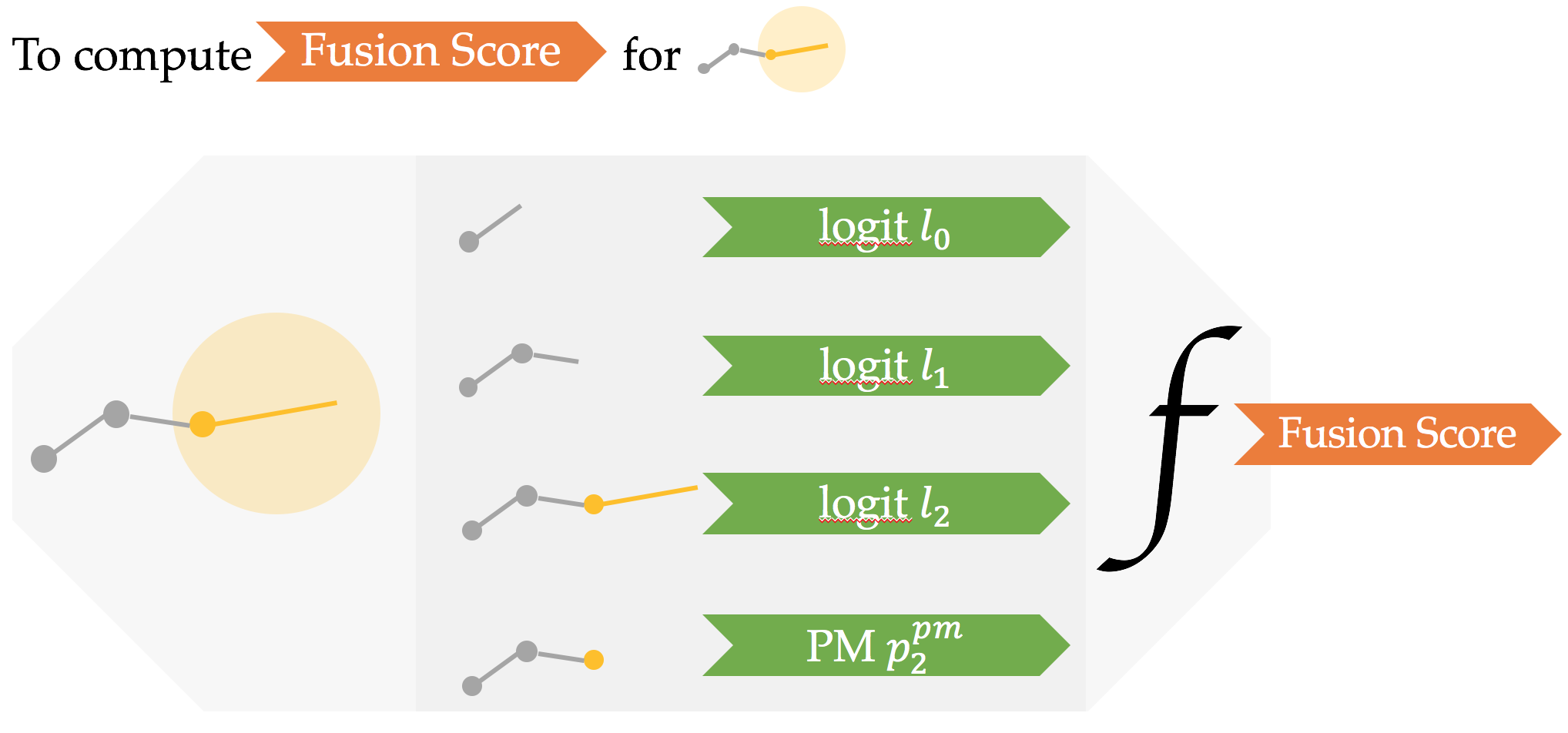}
}
\hspace{20pt}
\subfigure[An expansion queue maintains all possible next actions from all partial trajectories. The options are sorted by their scores (Figure \ref{fig:fast-fusion}) in order to select the next action.]{
\label{fig:fast-overall} 
\includegraphics[width=0.46\textwidth]{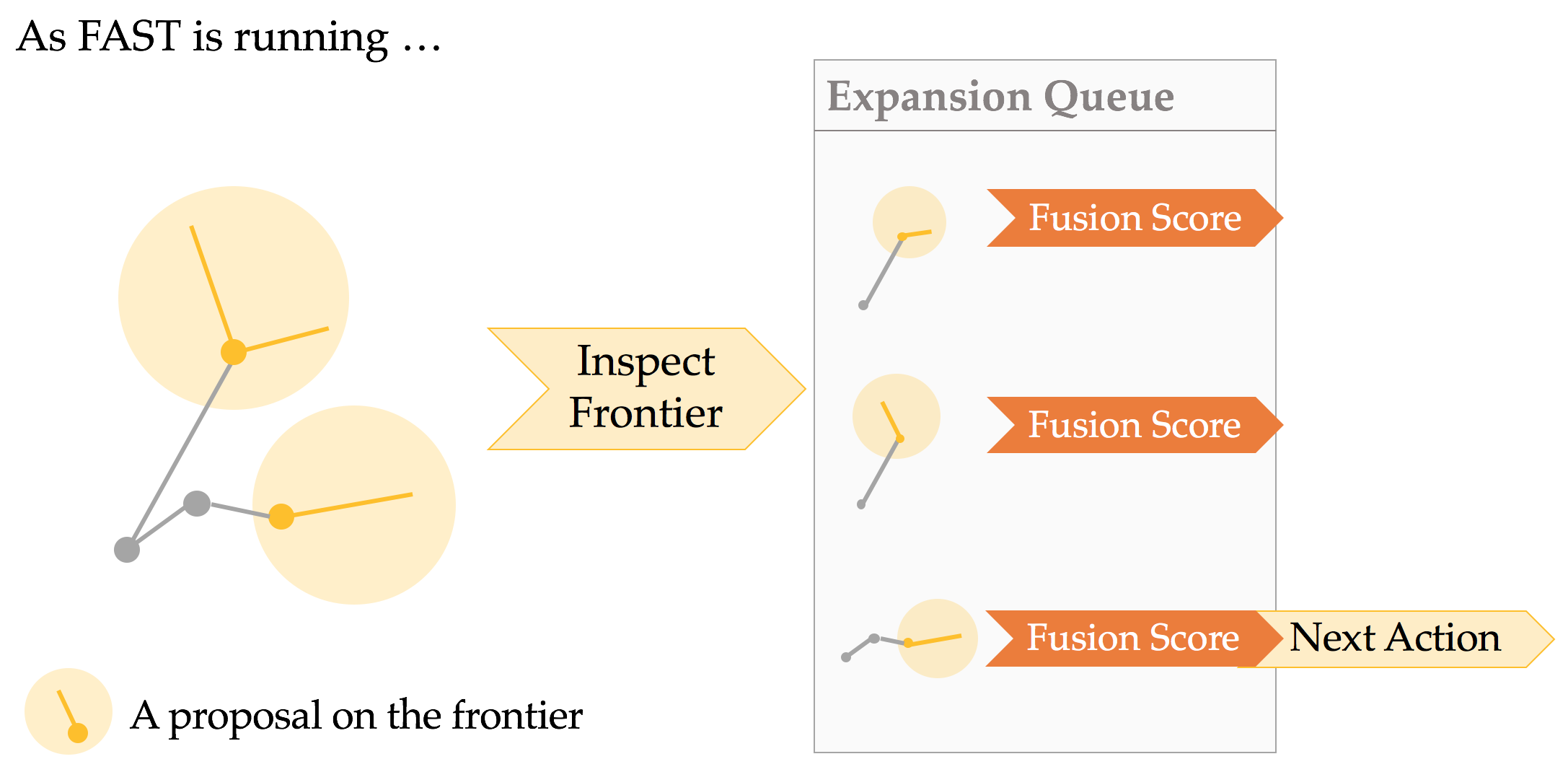}
}
\caption{Arbitrary signals can be computed from partial trajectories to learn a scoring function (left) that ranks all possible actions in our expansion queue (right). This provides a flexible and extendible framework for optimal action decoding.} 
\label{fig:fast-rerank} 
\end{figure*}

\paragraph{Q2: Where should we backtrack to?}
Making this decision involves using $\mathcal{L}$ to score all partial trajectories. Intuitively, the better a partial trajectory aligns with a given description, the higher the value of $\mathcal{L}$. Thus, if we can assume the veracity of $\mathcal{L}$, the agent simply returns to the highest scoring node when backtracking. Throughout this paper, we explore several functions for computing $\mathcal{L}$, but we present two simple techniques here, each acting over the sequence of actions that comprise a trajectory:
\begin{enumerate}[noitemsep,topsep=2pt,leftmargin=*]
    \item \textit{Sum-of-log $\sum_{0\rightarrow t} \log{p_i}$} sums the log-probabilities of every previous action, thereby computing the probability of a partial trajectory.
    \item \textit{Sum-of-logits $\sum_{0\rightarrow t} l_i$} sums the unnormalized logits of previous actions, which outperforms summing probabilities. These values are computed using an attention mechanism over the hidden state, observations, and language. In this way, their magnitude captures how well the action was aligned with the target description (this information is lost during normalization).\footnote{This is particularly problematic when an agent is lost. Normalizing many low-value logits can yield a comparatively high probability (e.g. uniform or random). We also experiment with variations of this approach (e.g. means instead of sums) in \S\ref{sec:analysis}.}\\
\end{enumerate}

Finally, during exploration, the agent implicitly constructs a ``mental map" of the visited space. This lets it search more efficient by refusing to revisit nodes, unless they lead to a high-value unexplored path.

\paragraph{Q3: Which visited node is most likely to be the goal?} Unlike existing approaches, \short{} considers every point that the agent has visited as a candidate for the final destination,\footnote{There can be more than one trajectory connecting the starting node to each visited node.} meaning we must rerank all candidates. We achieve this using $\mathcal{G}$, a trainable neural network function that incorporates all global information for each candidate and ranks them accordingly. Figure \ref{fig:fast-fusion} shows a simple visualization.

We experimented with several approaches to compute $\mathcal{G}$, e.g., by integrating $\mathcal{L}$, the progress monitor, speaker score, and a trainable ensemble in (\S\ref{sec:ablate_feats}).

\paragraph{Q4: When do we terminate the search?}
The flexibility of \short{} allows it to recover both the greedy decoding and beam search framework. In addition, we define two alternative stopping criteria:
\begin{enumerate}[noitemsep,topsep=2pt]
    \item When a partial trajectory decides to terminate.
    \item When we have expanded $M$ nodes. In \S\ref{sec:exp} we ablate the effect of choosing a different $M$.
\end{enumerate}

\subsection{Algorithm}
\label{sec:algorithm}
%\K{looks like we deleted the original 2.3 scoring subsection}
%To address the aforementioned questions, we maintain two queues: a frontier queue for (\XL{needs one sentence here}) and a global candidate queue for (\XL{needs one sentence here}). 
We present the algorithm flow of our \short{} framework.
When an agent is initialized and placed on the starting node, both the candidate and frontier queues are empty.
%\begin{eqnarray}
%    \mathcal{Q}_F && = \{\}\\
%    \mathcal{Q}_C && = \{\}\\
%\end{eqnarray}
The agent then adds all possible next actions to the frontier queue and adds its current location to the candidate queue:
\begin{eqnarray}
    Q_F && \leftarrow Q_F + \forall_{i\in K} \{\tau_0 \cup (u_0, a_i)\}\\
    Q_C && \leftarrow Q_C + \tau_0
\end{eqnarray}

Now that the $\mathcal{Q}_F$ is not empty and the stop criterion is not met, \short{} can choose the best partial trajectory from the frontier queue under the local scoring function:
\begin{eqnarray}
    \hat{\tau} && \leftarrow \arg\max_{\tau_i} \mathcal{L}(Q_F)
\end{eqnarray}
Following $\hat{\tau}$, we perform the final action proposal, $a_t$, to move to a new node (location in the house). \short{} can now update the candidate queue with this location and the frontier queue with all possible new actions. We then either continue, by exploiting the available actions at the new location, or backtrack, depending on the choice of backtrack criteria. We repeat this process until the model chooses to stop and returns the best candidate trajectory.  
\begin{eqnarray}
    \tau^* && \leftarrow \arg\max_{\tau} \mathcal{G}(Q_C)
\end{eqnarray}
Algorithm~\ref{alg:global_decoding} more precisely outlines the full procedure for our approach. \S\ref{sec:ablate_feats} details the different approaches to scoring partial and complete trajectories. 
%Pseudocode for our approach is presented in Algorithm \ref{alg:global_decoding}.

\begin{algorithm}[t]
  \caption{\decoder{}}
  \begin{algorithmic}[1]
    \Procedure{\decoder{}}{}
      \State $Q_F^{sort=\mathcal{L}}, Q_C^{sort=\mathcal{G}} = \{\}, \{\}$
      \State $Q_F \gets (u_0, a_0=\mathrm{\texttt{None}})$ \Comment{Initial Proposal}
      \State $\hat{\tau} \gets \varnothing$
      \State $M \gets \varnothing$ \Comment{Mental Map}
      \While{$Q_F \not= \varnothing$ and \textit{stop criterion}}
        \If{\textit{need backtrack} or $\hat{\tau} == \varnothing$}
            \State $\hat{\tau} \gets Q_F$.pop
        \EndIf
        \State $\hat{u}_{t-1}, \hat{a}_{t-1} \gets \hat{\tau}$.last
        
        \If{$(\hat{u}_{t-1}, \hat{a}_{t-1}) \in M$}
            \State $u_t \gets M(\hat{u}_{t-1}, \hat{a}_{t-1})$ 
        \Else
            \State $u_t \gets$ move to $u_{t-1}$ and execute $a_{t-1}$
            \State $M(\hat{u}_{t-1}, \hat{a}_{t-1}) \gets u_t$
        \EndIf
        %\State Rank actions following $u_{current}$ by $S_{local}$ 
        \For{$a_k$ in best $K$ next actions}
            \State $Q_F \gets Q_F \cup \{\hat{\tau} + (u_{t}, a_k)\}$
        \EndFor
        \State $Q_C \gets Q_C \cup \hat{\tau}$%, priority $H_{end}(\tau_{current})$
        \State $\hat{\tau} \gets \hat{\tau} + (u_{t}, a^*)$ where $a^*$ is the best action
      \EndWhile\label{euclidendwhile}
      \State \textbf{return} $Q_C$.pop
    \EndProcedure
  \end{algorithmic}
  \label{alg:global_decoding}
\end{algorithm}

\section{Experiments}
\label{sec:exp}

\label{sec:dataset}
We evaluate our approach using the Room-to-Room (R2R) dataset~\cite{anderson2018vision}. At the beginning of the task, the agent receives a natural language instruction and a specific start location in the environment; the agent must navigate to the target location specified in the instruction as quickly as possible. R2R is built upon the Matterport3D dataset~\cite{chang2017matterport3d}, which consists of $>$194K images, yielding 10,800 panoramic views (``nodes") and 7,189 paths. Each path is matched with three natural language instructions.
%The average physical path length is 10m and the average length of the instructions is 29 words. The whole dataset is split into four sets as detailed in Table~\ref{tab:dataset}: training, validation seen, validation unseen, and test unseen.

\subsection{Evaluation Criteria}
\label{sec:eval_metrics}
We evaluate our approach on the following metrics in the R2R dataset:
\begin{itemize}[noitemsep,topsep=2pt] %,leftmargin=*
    \item[\textbf{\texttt{TL}}] \textbf{Trajectory Length} measures the average length of the navigation trajectory.
    \item[\textbf{\texttt{NE}}] \textbf{Navigation Error} is the mean of the shortest path distance in meters between the agent's final location and the goal location. 
    \item[\textbf{\texttt{SR}}] \textbf{Success Rate} is the percentage of the agent's final location that is less than 3 meters away from the goal location.
    %\item Oracle Success Rate (OSR): is the success rate of the closest point on the agent's trajectory and the goal location, even if the agent does not stop there finally.
    \item[\textbf{\texttt{SPL}}] \textbf{Success weighted by Path Length~\cite{anderson2018evaluation}} trades-off \texttt{SR} against \texttt{TL}. Higher score represents more efficiency in navigation. 
\end{itemize}

\subsection{Baselines}
We compare our results to four published baselines for this task.\footnote{Some baselines on the leader-board are not yet public when submitted; therefore, we cannot compare with them directly on the training and validation sets.}
\begin{itemize}[noitemsep,topsep=0pt]
    \item \textsc{Random}: an agent that randomly selects a direction and moves five step in that direction ~\cite{anderson2018vision}. 
    \item \textsc{Seq2seq}: the best performing model in the R2R dataset paper~\cite{anderson2018vision}.
    %\item \textsc{RPA}~\cite{wang2018look}: is an agent which combines the model-free and model-based reinforcement learning, using a look-ahead module for planning. 
    \item \textsc{Speaker-Follower}~\cite{fried2018speaker}: an agent trained with data augmentation from a speaker model on the panoramic action space.
    \item \textsc{SMNA}~\cite{ma2019self}: an agent trained with a visual-textual co-grounding module and a progress monitor on the panoramic action space.\footnote{Our \textsc{SMNA} implementation matches published validation numbers. All our experiments are based on full re-implementations.}
    %\YB{Code is now available: https://github.com/chihyaoma/selfmonitoring-agent} but we could not reproduce their reported test numbers, It is therefore possible insights from their code will increase our performance by several points.
\end{itemize}

\subsection{Our Model}
% Here \short{} uses location information from the \textsc{SMNA} agent (baseline). 

As our framework provides a flexible design space, we report performance for two versions:
\begin{itemize}[noitemsep,topsep=2pt] %,leftmargin=*
    \item \short{}(short) uses the exploit strategy. We use the sum of logits \textit{fusion} method to compute $\mathcal{L}$ and  terminate when the best local action is stop.
    \item \short{}(long) uses the explore strategy. We again use the sum of logits for fusion, terminating the search after fixed number of nodes and using a trained neural network reranker to select the goal state $\mathcal{G}$.
\end{itemize}

%We reported our model's performance using local information from the \textsc{SMNA} agent. \short{} agents  \short{} (short) agent  to determine when to backtrack and to terminate the search when the best local action proposes to stop. In contrast, \short{} (long) agent , which incorporates into a single score the logits, the progress monitor of \textsc{SMNA} and the \textsc{Speaker} model from \textsc{Speaker-Follower}. This agent terminates after expanding $55$ nodes. 

\subsection{Results}
\label{sec:results}
\begin{table*}
\centering
%\small
\begin{tabular}{@{}l@{\hspace{3pt}}l@{}rccc|rccc|rccc@{}}%rrrrrrrrrrrr
\\ 
& & \multicolumn{4}{c}{Validation Seen} & \multicolumn{4}{c}{Validation Unseen} & \multicolumn{4}{c}{Test Unseen} \\ 
%\cline{2-13}
& Model & \texttt{TL} & \texttt{NE} & \texttt{SR} & \texttt{SPL} & \texttt{TL} & \texttt{NE} & \texttt{SR} & \texttt{SPL} & \texttt{TL} & \texttt{NE} & \texttt{SR} & \texttt{SPL} \\ 
\toprule
& \textsc{Random} & 9.58 & 9.45 & 0.16 & - & 9.77 & 9.23 & 0.16 & - & 9.93 & 9.77 & 0.13 & 0.12 \\
%\textsc{Shortest} & 10.19 & 0.00 & 1.00 & - & 9.48 & 0.00 & 1.00 & - & 9.93 & 0.00 & 1.00 & - \\
& Seq2seq & 11.33 & 6.01 & 0.39 & - & 8.39 & 7.81 & 0.22 & - & \phantom{0,0}8.13 & 7.85 & 0.20 & 0.18 \\
\rowcolor{Gray}
& Our baseline SMNA & 11.69 & 3.31 & 0.69 & 0.63 & 12.61  & 5.48 & 0.47 & 0.41 & - & - & - & - \\
\midrule
\multirow{2}{*}{\rotatebox{90}{\footnotesize Greedy}} & \textsc{SMNA} & - & - & - & - & - & - & - & - & \phantom{0,0}18.04 & 5.67 & 0.48 & 0.35 \\
& \textsc{Speaker-Follower} & - & - & - & - & - & - & - & - & \phantom{0,0}\textbf{14.82} & 6.62 & 0.35 & 0.28\\
\rowcolor{Gray}
& + \short~ (short) &  &  &  & & 21.17 & 4.97 & 0.56 & 0.43 & 22.08 & \textbf{5.14} & \textbf{0.54} & \textcolor{blue}{\textbf{0.41}}  \\
%RPA & - & 5.56 & 0.43 & - & - & 7.65 & 0.25 & - & 9.15 & 7.53 & 0.25 & 0.23 \\
\midrule
\multirow{2}{*}{\rotatebox{90}{\footnotesize Beam}} &\textsc{SMNA}  & - & 3.23 & 0.70 & - & - & 5.04 & 0.57 & - & \phantom{0,}373.09 & 4.48 & \textcolor{blue}{\textbf{0.61}} & 0.02 \\
& \textsc{Speaker-Follower}  & - & 3.88 & 0.63 & - & - & 5.24 & 0.50 & - & 1,257.30 & 4.87 & 0.53 & 0.01 \\
\rowcolor{Gray}
& + \short~ (long) & 188.06 & 3.13 & 0.70 & 0.04 & 224.42 & 4.03 & \textcolor{blue}{\textbf{0.63}} & 0.02 & \textbf{196.53} & \textbf{4.29} & \textcolor{blue}{\textbf{0.61}} & \textbf{0.03} \\
%& + \short~ (balanced) & 35.35 & 3.22 & 0.70 & 0.25 & 60.19 & 4.34 & 0.60 & 0.16 & 65.40 & 4.73 & 0.56 & 0.16 \\
\bottomrule
& Human & - & - & - & - & - & - & - & - & \phantom{0,0}11.85 & 1.61 & 0.86 & 0.76 \\
\bottomrule
\end{tabular}
\vspace{2mm}
\caption{Our results and SMNA re-implementation are shown in gray highlighted rows.  \textbf{Bolding} indicates the best value per section and \textcolor{blue}{\textbf{blue}} indicates best values overall.
We include both a short and long version of our approach to compare to existing models greedy and beam search approaches.%\K{note that I remove the "balanced" setting, it complicates the story and is not impressive on a single metrics}
}
\label{tab:main_result}
\end{table*}

Table \ref{tab:main_result} compares the performance of our model against published numbers of existing models. Our approach significantly outperforms the existing model in terms of efficiency, matching the best overall success rate despite taking 150 - 1,000 fewer steps. This efficiency gain can be seen in the \texttt{SPL} metric, where our models outperform previous approaches in every setting.
%\footnote{Section \S \ref{sec:analysis} presents ablation of these choices.} 
Note that our short trajectory model appreciably outperforms current approaches in both \texttt{SR} and \texttt{SPL}. %achieving a state-of-the-art of 0.41.  
If our agent could continue exploring, it matches existing peak success rates in half of the steps (196 vs 373).

\begin{table}[!ht]
    \centering
    \small
    \begin{tabular}{lr@{\hspace{3pt}}rr@{\hspace{3pt}}rc}%{@{}llll@{}}
        \toprule
        Validation Unseen & \multicolumn{2}{c}{\hspace{-10pt}\texttt{SR} (\%)} & \multicolumn{2}{c}{\hspace{-10pt}\texttt{SPL} (\%)} & \texttt{TL} \\
        \midrule
        \textsc{Speaker-Follower} & 37 & & 28 & & 15.32 \\
        \phantom{+} + \short{} & 43 & (\textcolor{blue}{\textbf{+6}}) & 29 & (\textcolor{blue}{\textbf{+1}}) & 20.63 \\
        \midrule
        \textsc{SMNA} & 47 & & 41 & & 12.61 \\
        \phantom{+} + \short{} & 56 & (\textcolor{blue}{\textbf{+9}}) & 43 & (\textcolor{blue}{\textbf{+2}}) & 21.17 \\
        \bottomrule
    \end{tabular}
    \vspace{2mm}
    \caption{Plug-n-play performance gains achieved by adding \short{} to current SoTA models.}
    \label{tab:res-g}
\end{table}

%\paragraph{Plug-n-Play} 
Another key advantage of our technique is how simple it is to integrate with current approaches to achieve dramatic performance gains. Table \ref{tab:res-g} shows how the sum-of-logits fusion method enhances the two previously best performing models. Simply changing their greedy decoders to \short{} with no added global information and therefore no reranking yields immediate gains of 6 and 9 points in success rate for \textsc{Speaker-Follower} and \textsc{SMNA}, respectively. Due to those models' new ability to backtrack, the trajectory lengths increase slightly. However, the success rate increases so much that SPL increases, as well.

%In the remainder of this paper we perform a larger comparison of both local and global scores and fusion heuristics. 

%\K{can we move/remove this?
%As noted in our methods sections and Algorithm~\ref{alg:global_decoding}, our approach provides several tunable parameters, including the ability to budget the agents actions. We demonstrate this, and the requisite stopping criteria, in section \ref{sec:know-to-stop}.}

%Our \short{} (short) achieves state-of-the-art \texttt {SPL}, and our (long) version is more efficient than its corresponding beam search models. \K{we might want to put short beneath greedy and long beneath beam in the table}
%($0.41 > 0.35/0.28$ and $0.03 > 0.02/0.01$). Our \short{}~(short) exploits the local information provider by trusting its decisions until the agent gets confused. In this instance, we interrupt the execution of greedy agent whenever it intends to move to a visited node on its partial trajectory. From there, the heuristic function steps in, reranks all partial trajectories on the frontier and puts the agent to a more desirable direction. 
%In addition, when measuring efficiency on validation unseen, SMNA states that with beam search they achieved 55\% success rate in 168.13 meters on validation unseen, which takes almost seven times as our \short{}~(short).\footnote{This result was stated in a response on OpenReview\\ \url{https://openreview.net/forum?id=r1GAsjC5Fm}}
%For \short~{}(long), we would expect our models \texttt{SR} to monotonically increase until converging with global information provider's performance as we increase our search budget. 

\section{Analysis}
\label{sec:analysis}

Here, we isolate the effects of local and global knowledge, the importance of backtracking, and various stopping criteria. In addition, we include three qualitative intuitive examples to illustrate the model's behavior in the Supplementary Materials (\S\ref{sec:app_examples}). We can perform this analysis because our approach has access to the same information as previous architectures, but it is more efficient.  Our claims and results are general, and our \short{} approach should benefit future VLN architectures.

\begin{figure}
\centerline{\includegraphics[width=0.9\columnwidth]{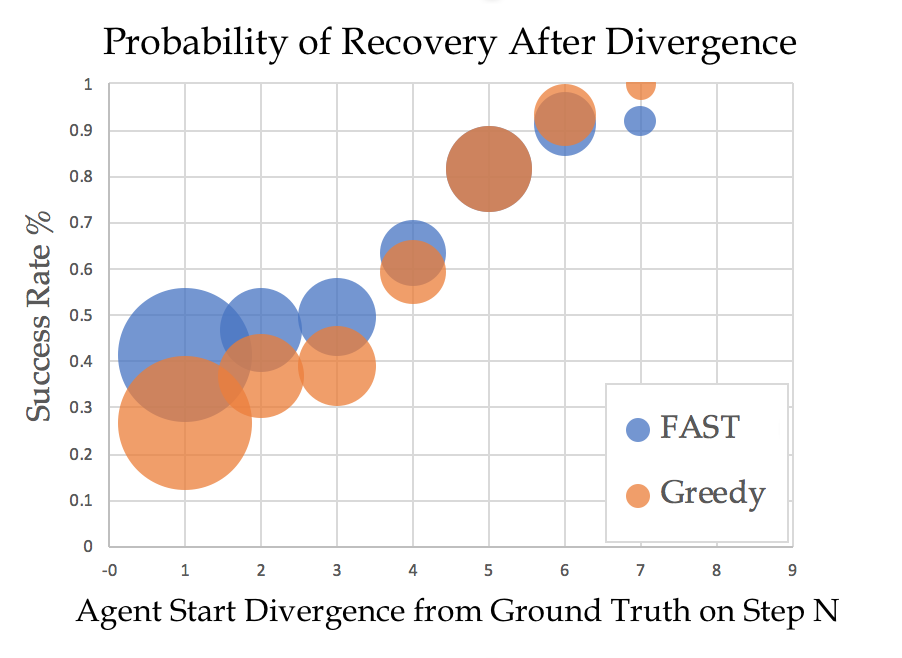}}%\quad
\caption{Circle sizes represent the what percentage of agents diverge on step N. Most divergences occur in the early steps. \short{} recovers from early divergences.}
\label{pics:recovery_from_divergenc}
\end{figure}

\subsection{Fixing Your Mistakes}
To investigate the degree to which models benefit from backtracking, Figure \ref{pics:recovery_from_divergenc} plots a model's likelihood of successfully completing the task after making its first mistake at each step. We use \textsc{SMNA} as our greedy baseline. Our analysis finds that the previous SoTA model makes a mistake at the very first action 40\% of the time. Figure \ref{pics:recovery_from_divergenc} shows the effect of this error: the greedy approach, if made a mistake at its first step, has a $<$30\% chance of successfully completing the task. In contrast, because \short{} detects its mistake, it returns to the starting position and tries again. This simple one-step backtracking increases its likelihood of success by over 10\%. In fact, the greedy approach is equally successful only if it progresses over halfway through the instruction without making a mistake.

\subsection{Knowing When To Stop Exploring}

\label{sec:know-to-stop}
\begin{figure}
\centering
\includegraphics[width=\columnwidth]{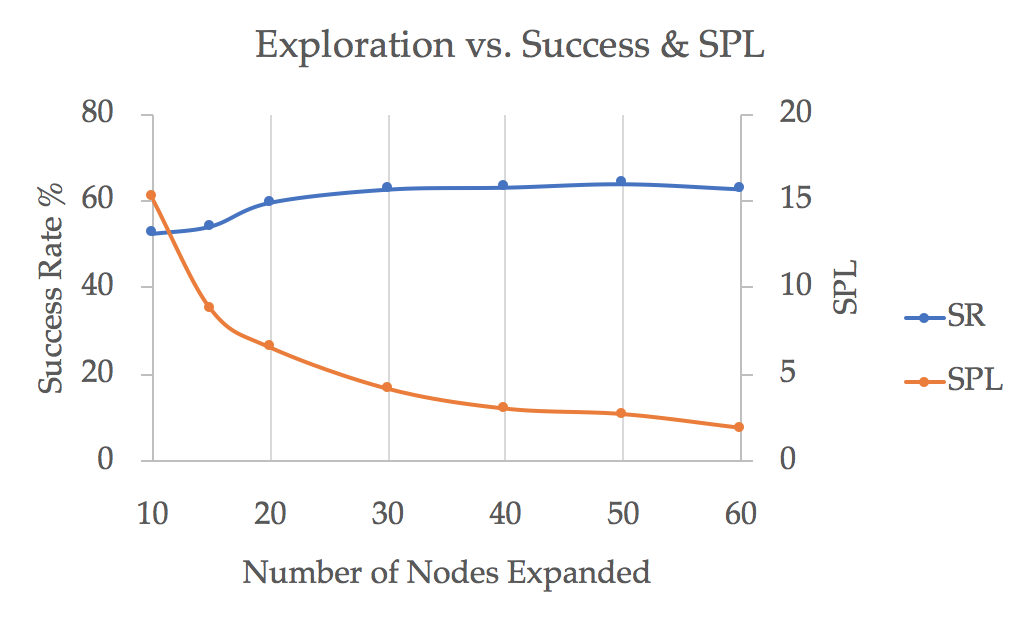}\quad
\caption{The SR increases with the number of nodes explored before plateauing, while SPL (which is extremely sensitive to length) continually decreases with added exploration.}
\label{fig:beam_pareto}
\end{figure}

The stopping criterion balances exploration and exploitation. Unlike previous approaches, our framework lets us compare different criteria and offers the flexibility to determine which is optimal for a given domain. The best available stopping criterion for VLN is not necessarily the best in general. We investigated the number of nodes to expand before terminating the algorithm, and we plot the resulting success rate and SPL in Figure \ref{fig:beam_pareto}. %With more nodes expanded, the model has a slight increase in success rate, while being less efficient and has lower SPL. 
One important finding is that the model's success rate, though increasing with more nodes expanded, does not match the oracle's rate, i.e., as the agent expands 40 nodes, it has visited the true target node over 90\% of the time but cannot recognize it as the final destination. This motivates an analysis of the utility of our global information and whether it is truly predictive (Table \ref{tab:heuristics}), which we investigate further in \S\ref{sec:ablate_feats}.
%\JG{Should we include the oracle rate in Figure~\ref{fig:beam_pareto}?}

%Because \texttt{SPL} penalizes according to trajectory length, it necessarily decreases with exploration (orange), despite success continuing to slowly increase as new paths are uncovered. \texttt{SR} performance however does not match the oracle. 

%\subsection{Using the Right Features}
\subsection{Local and Global Scoring}
\label{sec:ablate_feats}
%\K{this section requires some re-write, potentially combined with Sec 2.3. @yonatan}
As noted in \S\ref{sec:algorithm}, core to our approach are two queues, frontier queue for expansion and the candidate queue for proposing the final candidate. Each queue can use arbitrary information for scoring (partial) trajectories. We now compare the effects of combining different set of signals for scoring each queue.

\paragraph{Fusion methods for scoring partial trajectories}
An ideal model would include as much global information as possible when scoring partial trajectories in the frontier expansion queue. Thus, we investigated several sources of pseudo-global information and ten different ways to combine them. The first four use only local information, while the others attempts to fuse local and global information. 

The top half of Table~\ref{tab:res-logp-logit} shows the performance when considering only local information providers. For example, the third row of the table shows that summing the logit scores of nodes along the partial trajectory as the $\mathcal{L}$ score for that trajectory achieves an SR score of 56.66. Note although all information originates with the same hidden vectors, the values computed and how they are aggregated substantially affect performance. Overall, we find that summing unnormalized logits (the 3rd row) performs the best considering its outstanding SR. This suggests that important activation information in the network outputs is being thrown away by normalization and therefore discarded by other techniques.

The bottom part of Table~\ref{tab:res-logp-logit} explores ways of combining local and global information providers. These are motivated by beam-rescoring techniques in previous work (e.g., multiplying by the normalized progress monitor score).  Correctly integrating signals is challenging, in part due to differences in scale. For example, the logit is unbounded (+/-), log probabilities are unbounded in the negative, and the progress monitor is normalized to a score between 0 and 1. Unfortunately, direct integration of the progress monitor did not yield promising results, but future signals may prove more powerful.
% \JG{Why not use a trainable fusion function, similar to the MLP used for ranking completed trajectories, to combine all these features?}

\begin{table}
\centering
\vspace{-1mm}
\begin{tabular}{llccc}
%\toprule
%\multicolumn{5}{c}{\textbf{On Validation Unseen}} \\
\toprule
\textit{Heur/step} &  \textit{Combine} & \textit{SR} & \textit{SPL} &  \textit{Len} \\
\midrule
logit     & mean     & 53.89 & \textbf{44.74} & 14.80  \\
log prob  & mean     & 53.85 & 44.14 & 15.57  \\
logit     & sum      & \textbf{56.66} & 43.64 & 21.17 \\
log prob  & sum      & 56.23 & 42.66 & 21.70\\
\midrule
logit     & mean / pm     & 53.00 & 44.51 & 13.67 \\
log prob  & mean / pm    &  53.72 & 44.64 & 13.85 \\
logit     & mean * pm    &  54.78 & \textbf{44.70} & 15.91 \\
log prob  & mean * pm    &  55.04 & 43.70 & 17.45 \\
logit     & sum * pm    &  50.95 & 41.28 & 20.25 \\
log prob  & sum * pm    &  \textbf{56.15} & 43.19 & 21.55 \\
\bottomrule
\end{tabular}
\vspace{2mm}
\caption{Performance of different fusion methods for scoring partial trajectories. Tested on the validation unseen set.}
\label{tab:res-logp-logit}
\vspace{-2mm}
\end{table}

\paragraph{Fusion methods for ranking complete trajectories}.
Previous work~\cite{fried2018speaker} used state-factored beam search to generate $M$ candidates and rank the complete trajectories using probability of speaker and follower scores $\text{argmax}_{r\in R(d)} P_S(d|r)^\lambda * P_F(d|r)^{(1-\lambda)}$. In addition to the speaker and progress monitor scores used by previous models, we also experiment with using $\mathcal{L}$ to compute $\mathcal{G}$.
%In addition to the information above, we also experiment with the speaker score from previous work as a global signal. 
To inspect the performance of using different fusion methods, we ran \decoder{} to expand 40 nodes on the frontier and collect candidate trajectories. Table~\ref{tab:heuristics} shows the performance of different fusion scores that rank complete trajectories. We see that most techniques have a limited understanding of the global task's goal and formulation. We do, however, find a significant improvement on unseen trajectories when all signals are combined. For this we train a multi-layer perceptron to aggregate and weight our predictors. Note that any improvements to the underlying models or new features introduced by future work will directly correlate to gains in this component of the pipeline.

The top line of Table~\ref{tab:heuristics}, shows oracle's performance.  This indicates how far current global information providers have yet to achieve. Closing this gap is an important direction for future work.

\begin{table}[ht!]
\centering
\begin{tabular}{lccc}
\toprule
% & \textbf{Train}  &  \textbf{Val Seen} & \textbf{Val Unseen}\\
 & Train & Val Seen & Val Unseen \\
\midrule
\textit{Oracle} & 99.13 & 92.85 & 90.20 \\
\midrule
\textit{$\Sigma\  l_i$}    & 78.78 & 62.49 & 56.49\\
\textit{$\mu\  l_i$}    & 85.78 & 66.99 & 54.41\\
\textit{$\Sigma\  p_i$}   & 91.25 & 68.56 & 56.15\\
\textit{$\mu\  p_i$}   & 91.60 & 69.34 & 58.75\\
\textit{$p^{pm}_t$}   & 66.71 & 53.67 & 50.15\\
\textit{$\mathcal{S}$}    & 69.99 & 53.77 & 43.68\\
\midrule
\textit{All}   & 90.16 & \textcolor{blue}{\textbf{71.00}} & \textcolor{blue}{\textbf{64.03}}\\
\bottomrule
\end{tabular}
\vspace{2mm}
\caption{Success rate using seven different fusion scores as $\mathcal{G}$ to rerank the destination node from the candidate pool. 
%This table tests performance of each piece of global information for reranker.
}
\label{tab:heuristics}
\end{table}

\subsection{Intuitive Behavior}
%\subsection{Qualitative comparison}
The Supplementary Materials (\S\ref{sec:supp_examples}) provide three real examples to show how our model performs when compared to greedy decoding (SMNA model). It highlights how the same observations can lead to drastically different behaviors during an agent's rollout. Specifically, in Figures~\ref{fig:supp_example1} and ~\ref{fig:supp_example2}, the greedy decoder is forced into a behavioral loop because only local improvements are considered. Using \short{} clearly shows that even a single backtracking step can free the agent of poor behavioral choices.

%Finally, we provide a simple qualitative example (in Figure~\ref{pics:example}) to show how our model performs when compared to greedy decoding. It highlights how the same observations leads to drastically different behavior during an agent's rollout. Specifically, the greedy decoder is forced to continue down a sub-optimal path. Confused by its path, the greedy decoder can be forced into a behavioral loop, because only local improvements are considered. Search makes it clear that even a single backtracking step can break the agent out of poor behavioral choices. In the Supplementary Material, we highlights the distinct behaviors of our model along side the greedy decoding scheme. 
\section{Related Work}
\label{sec:related}

Our work focuses on and complements recent advances in Vision-and-Language Navigation (VLN) as introduced by \cite{anderson2018vision}, but many aspects of the task and core technologies date back much further. The natural language community has explored instruction following using 2D maps \cite{misra2017mapping, mei2016listen} and computer-rendered 3D environments \cite{misra2018mapping}. Due to the enormous visual complexity of real-world scenes, the VLN literature usually builds on computer vision work from referring expressions \cite{mirowski2016learning,zhu2017target}, visual question answering \cite{antol2015vqa}, and ego-centric QA that requires navigation to answer questions \cite{gordon2017iqa,das2018embodied,de2018talk}. Finally, core to the our work is the field of search algorithm, dating back to the earliest days of AI~\cite{pearl1984heuristics,russell2016artificial}, but largely absent from recent VLN literature that tends to focuses more on neural architecture design.

During publishing the Room-to-Room dataset (VLN), \cite{anderson2018vision} introduced the ``student forcing" method for seq2seq model. Later work integrated a planning module to combined model-based and model-free reinforcement learning to better generalize to unseen environments~\cite{wang2018look}, and a Cross-Modal Matching method that enforces cross-modal grounding both locally and globally via reinforcement learning~\cite{wang2018reinforced}. Two substantial improvements came from panoramic action spaces and a ``speaker" model trained to enable data augmentation and trajectory reranking for beam search~\cite{fried2018speaker}. Most recently, \cite{ma2019self} leverages a visual-textual co-grounding attention mechanism to better align the instruction and visual scenes and incorporates a progress monitor to estimate the agent's current progress towards a goal. These approaches require beam search for peak SR. Beam search techniques can unfortunately lead to long trajectories when exploring unknown environments. This limitation motivates the work we present here. Existing approaches trade off a high success rate and long trajectories: greedy decoding provides short, often incorrect paths, the beam search yields high success rates but long trajectories. 
% Our goal here is to show that, rather than solving the same problems as end-to-end neural models, techniques from search can be used to harness the full potential of neural systems in ways that greedy decoding and beam search fail to. 
\section{Conclusion}
We present \decoder{}, a framework for using asynchronous search to boost any VLN navigator by enabling explicit backtrack when an agent detects if it is lost. This framework can be easily plugged into the most advanced agents to immediately improve their efficiency. Further, empirical results on the Room-to-Room dataset show that our agent achieves state-of-the-art Success Rates and SPLs. Our search-based method is easily extendible to more challenging settings, e.g., when an agent is given a goal without any route instruction~\cite{chaplot2017gated,hermann2017grounded}, or a complicated real visual environment~\cite{chen2018touchdown}.

\section*{Acknowledgments}
Partial funding provided by DARPA's CwC program through ARO (W911NF-15-1-0543), NSF (IIS-1524371, 1703166), National Institute of Health (R01EB019335), National Science Foundation CPS (1544797), National Science Foundation NRI (1637748), the Office of Naval Research, the RCTA, Amazon, and Honda.
%Additionally, We would like to thank the anonymous reviewers, members of the xlab at the University of Washington for their insightful comments on the work. Part of this work was done while LK was interning with Microsoft Research.

{\small
\bibliographystyle{ieee}
\bibliography{egbib}
}

%\newpage~
\newpage
\beginsupplement
\section{Supplementary Material}
Our appendix is structured to provide both corresponding qualitative examples for the quantitative results in the paper and additional implementation details for replication.

\subsection{Qualitative comparison}
\label{sec:app_examples}
\begin{figure*}[!b]
\vspace{-1mm}
\includegraphics[width=0.9\textwidth]{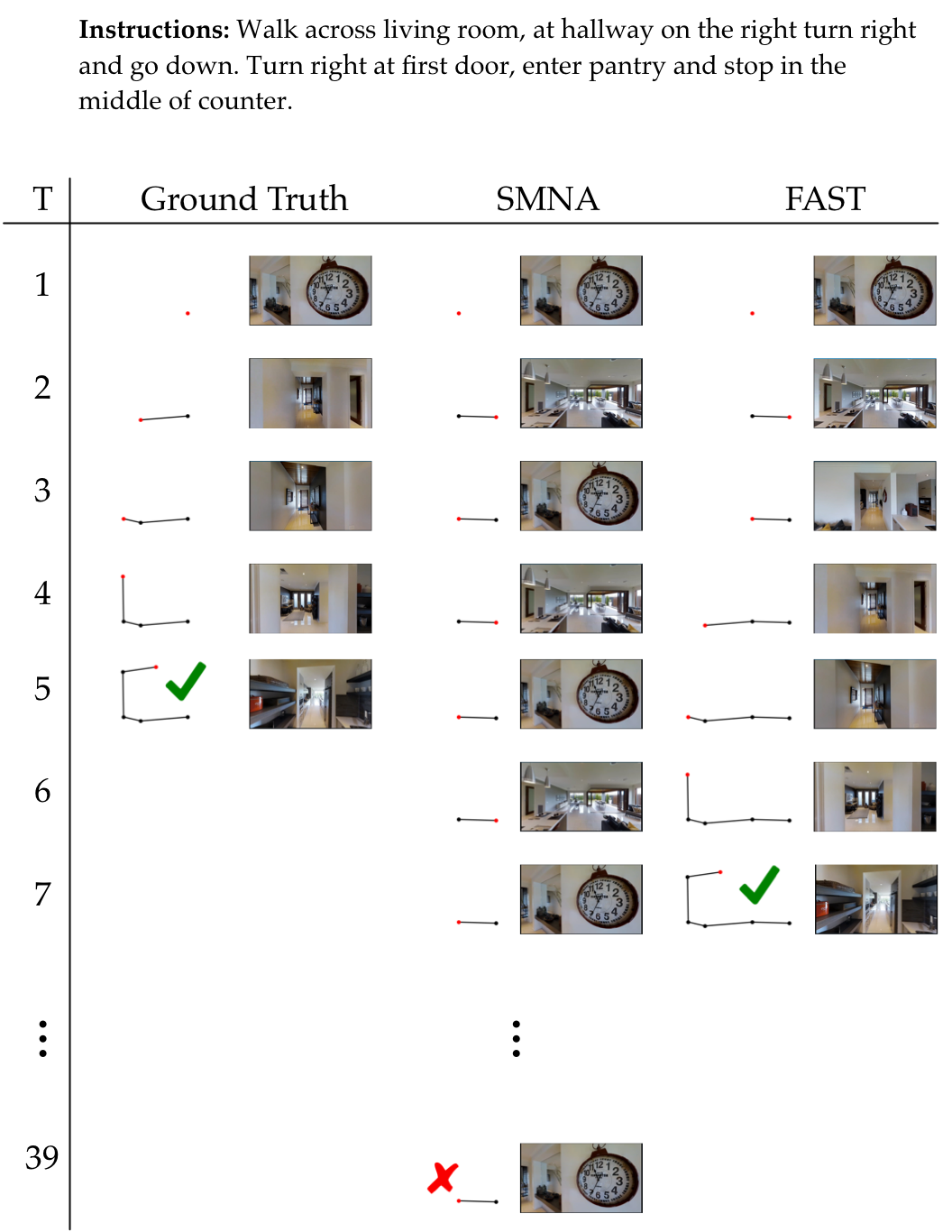}
\caption{Comparison of the previously state-of-the-art SMNA model \cite{ma2019self} to our \decoder{} method, with the ground truth as reference. Note how SMNA retraces its steps multiple times due to the lack of global information. This example is taken from Room-to-Room, path 2617, instruction set 3. You can view a video of this trajectory here: \url{https://youtu.be/AD9TNohXoPA}.}
\label{fig:supp_example1}
\end{figure*}

\begin{figure*}[!t]
\includegraphics[width=0.9\textwidth]{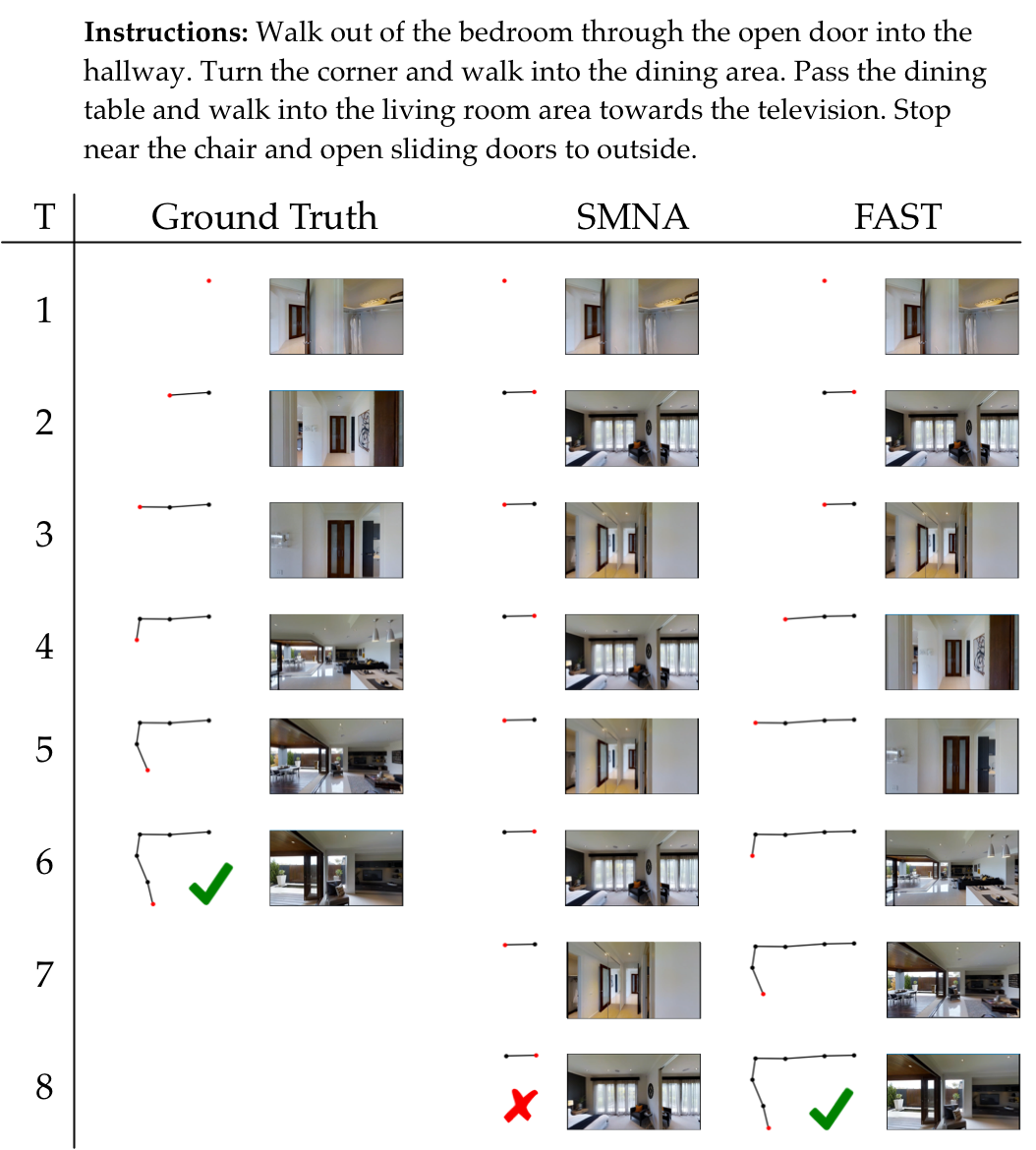}
\caption{Identical to previous figure \ref{fig:supp_example1}, except that this example is taken from Room-to-Room, path 15, instruction set 1. }
\label{fig:supp_example2}
\end{figure*}

\begin{figure*}
\includegraphics[width=0.9\textwidth]{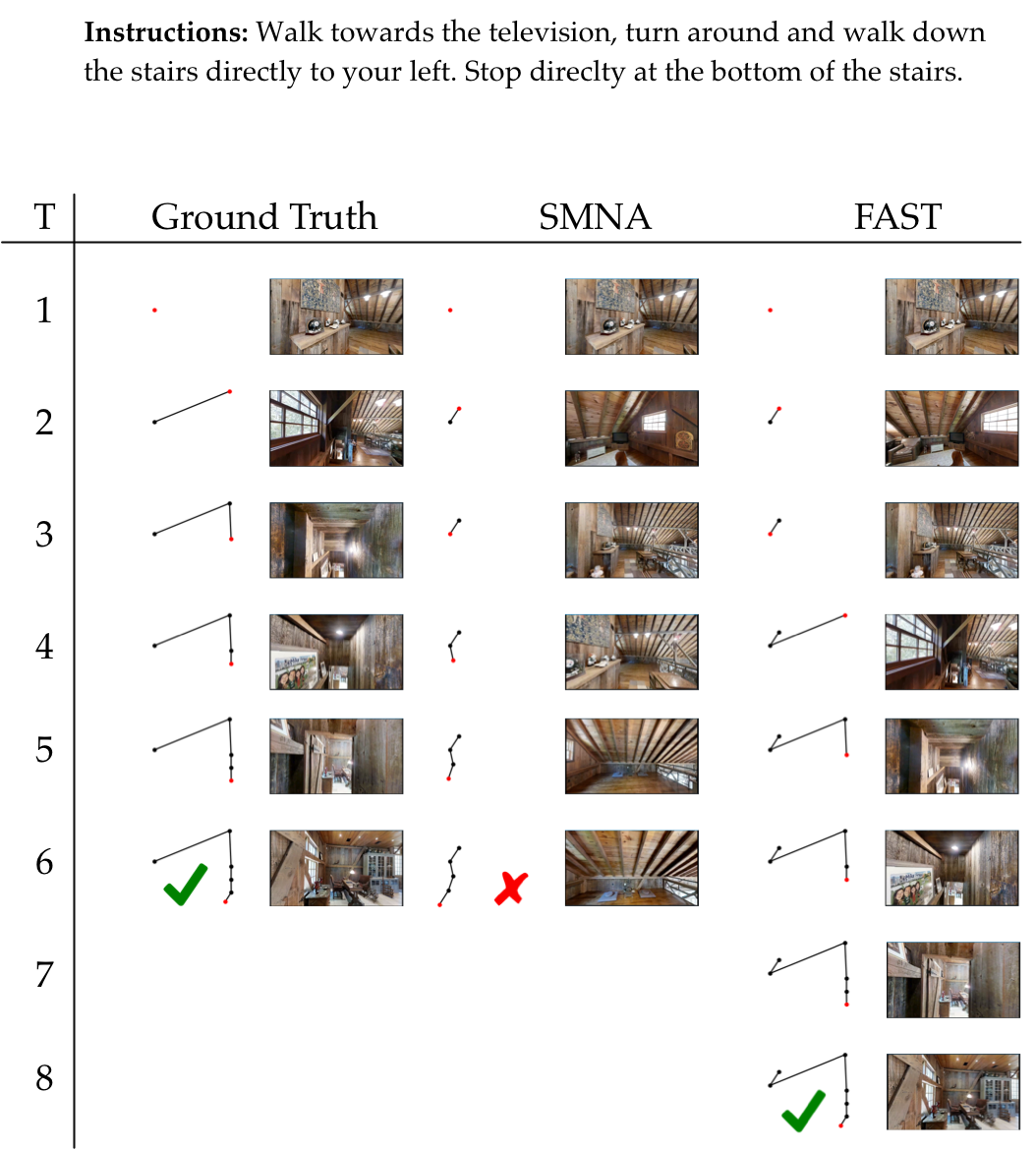}
\caption{Identical to previous figure \ref{fig:supp_example1}, except that this example is taken from Room-to-Room, path 1759, instruction set 1. The typo 'direclty' comes from the dataset.}
\label{fig:supp_example3}
\setlength{\belowcaptionskip}{10pt}
\end{figure*}

Figures~\ref{fig:supp_example1} through \ref{fig:supp_example3} show three examples comparing our approach to the previous state-of-the-art. In addition, the following URL includes a 90 second video (\url{https://youtu.be/AD9TNohXoPA}) showing a first-person view of several agents navigating the environment with corresponding birds-eye-view maps.
%\url{https://www.dropbox.com/s/uhhyj5owta0zi4a} 
\label{sec:supp_examples}

\subsection{Candidate Reranker}

Given a collection of candidate trajectories, our reranker module assigns a score to each of the trajectories. The highest scoring trajectory is selected for the \short{} agent's next step. In our implementation, we use a 2-layer MLP as the reranker. We train the neural network using pairwise cross-entropy loss~\cite{burges2005learning}.

As input to the reranker, we concatenate the following features to obtain a 6-dimensional vector:
\begin{itemize}[noitemsep]
    \item Sum of score logits for actions on the trajectory.
    \item Mean of score logits for actions on the trajectory.
    \item Sum of log probabilities for actions on the trajectory.
    \item Mean of log probability for actions on the trajectory.
    \item Progress monitor score for the completed trajector.
    \item Speaker score for the completed trajectory.
\end{itemize}

We feed the 6-dimensional vector through an \textit{MLP}: \texttt{BN $\rightarrow$ FC $\rightarrow$ BN $\rightarrow$ Tanh $\rightarrow$ FC}, where \texttt{BN} is a layer of \texttt{Batch Normalization}, \texttt{FC} is a \texttt{Fully Connected} layer, and \texttt{Tanh} is the nonlinearity used. The first \texttt{FC} layer transforms the 6-dimensional input vector to a 6-dimensional hidden vector. The second \texttt{FC} layer project the 6-dimensional vector to a single floating-point value, which is used as the score for the given partial trajectory. 

To train the \textit{MLP}, we cache the candidate queue after running \short{} for 40 steps. Each candidate trajectory in the queue has a corresponding score $s_i$. To calculate the loss, we minimize the pairwise cross-entropy loss:
$$-(s_1-s_2) + \log(1 + \exp(s_1-s_2))$$
where $s_1$ is the score for a qualified candidate and $s_2$ is the score for an unqualified candidate. We define \textit{qualified candidate trajectories} as those that end within 3 meters of ground truth destination. In our cached training set, we have $4,378,729$ pairs of training data. We train using a batch size of $3600$, SGD optimizer with a learning rate of  $5e^{-5}$, and momentum $0.6$; We train for $30$ epochs.

\clearpage

\clearpage

\end{document}